\documentclass{article}

     \usepackage[preprint]{neurips_data_2023}

\usepackage[utf8]{inputenc} 
\usepackage[T1]{fontenc}    
\usepackage{hyperref}       
\usepackage{url}            
\usepackage{booktabs}       
\usepackage{amsfonts}       
\usepackage{nicefrac}       
\usepackage{microtype}      
\usepackage{xcolor}         

\definecolor{codegreen}{rgb}{0,0.6,0}
\definecolor{codegray}{rgb}{0.5,0.5,0.5}
\definecolor{codepurple}{rgb}{0.58,0,0.82}
\definecolor{backcolour}{rgb}{0.95,0.95,0.92}

\usepackage{listings}
\lstdefinestyle{mystyle}{
    backgroundcolor=\color{backcolour},   
    commentstyle=\color{codegreen},
    keywordstyle=\color{magenta},
    numberstyle=\tiny\color{codegray},
    stringstyle=\color{codepurple},
    basicstyle=\ttfamily\footnotesize,
    breakatwhitespace=false,         
    breaklines=true,                 
    captionpos=b,                    
    keepspaces=true,                 
    numbers=left,                    
    numbersep=5pt,                  
    showspaces=false,                
    showstringspaces=false,
    showtabs=false,                  
    tabsize=2
}
\usepackage{amssymb}
\usepackage{amsthm}
\usepackage{caption}
\usepackage{subcaption}
\usepackage{graphicx}
\usepackage[symbol]{footmisc}
\usepackage{cancel}
\hypersetup{colorlinks,citecolor=blue,linkcolor=red,urlcolor=blue}
\usepackage{mathtools}
\usepackage{tabularx}
\usepackage{booktabs} 

\theoremstyle{definition}

\newcommand{\nocontentsline}[3]{}
\newcommand{\tocless}[2]{\bgroup\let\addcontentsline=\nocontentsline#1{#2}\egroup}

\title{Europepolls: A Dataset of Country-Level Opinion Polling Data for the European Union and the UK}

\author{%
  Konstantinos Pitas\\
  Statify Team\\
  INRIA Grenoble Rhône-Alpes\\
  Grenolbe, France \\
  \texttt{pitas.konstantinos@inria.fr} \\
}

\begin{document}

\maketitle

\begin{abstract}
I propose an open dataset of country-level historical opinion polling data for the European Union and the UK. The dataset aims to fill a gap in available opinion polling data for the European Union. Some existing datasets are restricted to the past five years, limiting research opportunities. At the same time, some larger proprietary datasets exist but are available only in a visual preprocessed time series format. Finally, while other large datasets for individual countries might exist, these could be inaccessible due to language barriers. The data was gathered from Wikipedia \citep{wikipedia}, and preprocessed using the pandas library \citep{mckinney-proc-scipy-2010}. Both the raw and the preprocessed data are in the .csv format. I hope that given the recent advances in LLMs and deep learning in general, this large dataset will enable researchers to uncover complex interactions between multimodal data (news articles, economic indicators, social media) and voting behavior. The raw data, the preprocessed data, and the preprocessing scripts are available on GitHub \href{https://github.com/konstantinos-p/europepolls}{konstantinos-p/europepolls}.
\end{abstract}

\tocless{\section{Introduction}}
Election results have significant implications for the citizens of a democratic country. However, the forecasting of election outcomes using statistical methods (a field known as ``psephology") has generally had a poor reputation \citep{jennings2020election}. This is in stark contrast to the forecasting of aggregate economic variables, stocks, and indeed many other types of highly complex data (from vision to audio and text) which machine learning has tackled with huge success in the past decades \citep{lecun2015deep}. However, researchers trying to forecast elections face a number of unique challenges \citep{silver2012signal,hummel2013fundamental}. The main challenge is a lack of available data simply by the very nature of the democratic process \citep{silver2012signal}. For example, the country of Spain has only held 16 elections since the restoration of democracy, and in fact, it has only held 16 free and fair elections since 1936 (the outbreak of the Spanish civil war). This creates a significant dearth of training data for any statistical model, let alone a modern deep learning model. Second, elections should be in principle affected by the entirety of the civic and economic life of a country in the timespan between two elections. This creates a highly multimodal regression problem, with (probably) significant non-linearities. 

I propose the Europepolls dataset, a dataset of opinion polling data for European Union countries. The proposed dataset has a number of favorable properties

\begin{itemize}
    \item The Europepolls dataset is open-source. The raw and preprocessed data as well as the preprocessing algorithm have been made public on GitHub under the CC BY-NC 4.0 (non-commercial) license. This presents significant opportunities for researchers, as other large datasets are proprietary and can only be accessed through an inferred time series.
    \item The opinion polling data cover the entire European Union, and have been standardized to the best of my ability. This increases the dataset size significantly to 120 elections and allows for exploring and exploiting inter-country correlations. Having such a standardized dataset is also a significant asset in a highly multilingual entity such as the European Union, where valuable data could be hidden behind language barriers.
    \item The opinion polling data cover the largest time span possible. Polls for different countries across the same time period should be to some extent correlated. Increasing the time-span of collected polls allows to include tail events such as economic crises.
\end{itemize}

The proposed dataset should be a valuable asset in the forecasting of elections and the understanding of voter behavior. A research question that I think would be valuable to explore is the extent to which news in the form of video, audio, and text affect user behavior. Indeed election forecasting models incorporate typically either prediction markets, opinion polling or aggregate economic data \citep{lewis2005election} but not more complex video, audio, and text data. While previous machine learning approaches allowed for the extraction of simple features such as sentiment features from such data perhaps limiting it's effectiveness, modern LLMs and ConvNets should be able to extract much more nuanced features. These features could then be used to find correlations with a time series inferred from the Europepolls dataset, and thus voting intentions.

I conduct some simple experiments that highlight the potential of the Europepolls dataset. Specifically, I estimate the systematic bias of each pollster towards each party (also known as the ``house effects"), for different countries and elections. Using this simple analysis one can find i) the evolution of polling bias across different elections ii) how different European Union countries compare in the bias of their opinion polls. Using such an analysis one can debias existing polls, resulting in a ``clean" time series that can form the basis for more involved analyses.

\tocless{\section{Related Work}}
Several entities systematically aggregate opinion polling data for the European Union. \href{https://europeelects.eu/our-data/}{EuropeElects} is a polling aggregation website run by a non-partisan team of 40 volunteers. EuropeElects volunteers gather polling data from their respective countries, clean them and then post the clean and open datasets on the EuropeElects website. A key limitation of the EuropeElects dataset is that data is typically restricted to 2018 and later. \href{https://www.politico.eu/europe-poll-of-polls/greece/}{Politico Poll-of-Polls} is a polling aggregator that gathers polling data from across the European Union. 
The data of Poll-of-Polls typically covers a much larger time span than EuropeElects data. However, the data come with severe restrictions. In particular, only a visual representation of the data in time-series form is available. It is not possible to download the data, even in this time-series form. Similarly \href{https://politpro.eu/en}{PolitPro} is another polling aggregation website. Similar restrictions exist as with Politico Poll-of-polls. The data is only available in time-series form, and even this cannot be downloaded.
 Various other datasets exist for independent countries. For example \href{https://www.markpack.org.uk/opinion-polls/}{PollBase} and \href{https://github.com/jackobailey/britpol}{Britpol} gather UK polling data, with the first going as far back as 1943. A key limitation of these datasets is that they do not allow an easy overview of historical voting trends across the EU.

\tocless{\section{Collection Method}}
The polling data for the different European countries is available per election in table form across various different pages of Wikipedia. I downloaded the raw table data in csv form using the \href{https://wikitable2csv.ggor.de/}{wikitable2csv} online tool. During this initial collection phase, I only applied very minor modifications to the downloaded csv files, mainly deleting some spurious rows that sometimes resulted from the automatic scrapping of the tables.

\tocless{\section{Format}}
Polling data are organized by election year. Thus the folder europepolls/countries/Austria/ contains the files 2013-09-29\_to\_2017-10-15, 2017-10-15\_to\_2019-09-29, 2019-09-29\_to\_2023-09-29 which correspond to three electoral periods. Each electoral folder contains a preprocessed\_polls.csv file which contains the preprocessed data. The electoral folder also contains a ``data" subfolder which contains .csv files that correspond to the raw polls scraped from Wikipedia.

Each preprocessed\_polls.csv file has the following format.

\begin{center}
\begin{tabular}{ |c|c|c|c|c|c| } 
 \hline
 Date & Polling Firm & Commissioner & Sample Size & Named Parties & Others \\ 
 \hline
 pandas.datetime & str & str & float & float & float \\ 
 \hline
\end{tabular}
\end{center}

A brief explanation of each column is given below.
\begin{itemize}
\item \textbf{Date}: The date when the poll was conducted. In some cases both a start and an end date is reported in the raw data,
corresponding to the start and the end of the polling. In this case, we report the end date. This column has ``pandas.datetime" entries.
\item \textbf{Polling Firm}: The name of the firm conducting the poll. This column has ``str" entries.
\item \textbf{Commissioner}: The name of the public or private entity which commissioned the poll. This column has ``str" entries.
\item Sample Size: The size of the sample of the poll. This column has ``float" entries.
\item \textbf{Named Parties}: The polling percentage of all parties with high enough polling numbers such that they are included by name in the poll. This column has ``float" entries ranging from 0-100\%. 
\item \textbf{Others}: This entry includes the polling total of all parties polling too low to be included by name in the results.
\end{itemize}

\tocless{\section{Preprocessing}}
Organizing voting-intention polling data presents a number of difficulties. Some of these difficulties stem from the inherent factors of the political systems of the European Union member countries. Other difficulties stem from the specific form in which the data was organized on the various Wikipedia pages.

\tocless{\subsection{Inherent difficulties of polling data.}}
I now describe the inherent difficulties in organizing a dataset of opinion poll data across different countries and different electoral cycles.
\begin{itemize}
    \item New political parties constantly emerge, and existing political parties dissolve. This often involves the renaming of different parties across the same electoral period. This creates the question of how these parties should be represented in a table of data. By and large, the convention that I followed is that when a party was renamed during the electoral cycle it appears as a single column with the name that it had during the election.
    \item Political parties merge into coalitions, and coalitions break up. Here the question that arises is whether the coalition should be listed as a single entry, multiple entries (one for each individual party), or both. Typically, the convention that I followed is that I list only the individual parties that constitute the coalition. Thus in subsequent analyses, one would need to know that a coalition had occurred so as to estimate the combined polling percentages of that coalition.
    \item Some countries (for example Belgium) have separate parties and separate polls for different regions (Wallonia and Flanders). Here the convention that I followed is that I aggregated all parties in the same table. 
    \item Some countries e.x. France has two elections that might be of interest the presidential election and the legislative election. Here I opted to present only the results of the presidential election, which is also the most consequential.
\end{itemize}

\begin{table}[!t]
  \caption{The first and final year for which opinion polls are recorded in the Europepolls dataset. A dash - signifies that polls are unavailable or of poor quality.}
  \label{table-start-end-date}
  \centering
  \begin{tabular}{lll}
    \toprule
 Country        & First Year & Final Year \\    
\midrule
 Austria        & 2013       & 2022       \\
 Belgium        & -          & 2022       \\
 Bulgaria       & 2014       & 2022       \\
 Croatia        & 2011       & 2022       \\
 Cyprus         & 2008       & 2022       \\
 Czech Republic & 2013       & 2022       \\
 Denmark        & 2011       & 2022       \\
 Estonia        & 2011       & 2022       \\
 Finland        & 2011       & 2022       \\
 France         & 2007       & 2022       \\
 Germany        & 2009       & 2022       \\
 Greece         & 2004       & 2022       \\
 Hungary        & 2006       & 2022       \\
 Ireland        & 2007       & 2022       \\
 Italy          & 2006       & 2022       \\
 Latvia         & 2014       & 2022       \\
 Lithuania      & 2016       & 2022       \\
 Luxembourg     & -          & 2022       \\
 Malta          & 2013       & 2022       \\
 Netherlands    & -          & 2022       \\
 Norway         & 2013       & 2022       \\
 Poland         & 1991       & 2022       \\
 Portugal       & 1999       & 2022       \\
 Romania        & 2012       & 2022       \\
 Slovakia       & 2012       & 2022       \\
 Slovenia       & 2008       & 2022       \\
 Spain          & 1986       & 2022       \\
 Sweden         & 2006       & 2022       \\
 Switzerland    & -          & 2022       \\
 UK             & 1983       & 2022       \\
    \bottomrule
  \end{tabular}
\end{table}

\tocless{\subsection{Difficulties with the Wikipedia data.}}
There are also a number of difficulties stemming from how opinion polling data is stored on Wikipedia. Primarily these are the result of the data not having been curated by a single central entity.
\begin{itemize}
\item Field names are not standardized. For example, the entity that paid for the poll might be referred to by different names such as ``Commissioner", ``Ordered by" etc. 
\item Undecided Voters/Blank Votes. These are voters that refuse to answer to the polls or state that they would cast a blank vote (in some jurisdictions considered a spoilt vote). In most cases, the results stated in Wikipedia data files have been renormalized such that they reflect polling percentages over \emph{valid votes} thus excluding Undecided Voters/ Blank Votes. In some cases though Undecided Voters/ Blank Votes are included as a separate column. In this case, I removed the Undecided Voters/ Blank Votes columns and renormalized the rest of the columns such that they summed up to 100\% of the vote.
\item Some fields are only included in some countries. For example ``Turnout" is a field recorded for some elections, and as a consequence appears in polling data csv files as well. Such spurious fields need to be removed.
\item In some cases the anonymous contributors to the opinion polling data have added superfluous information to the poll tables that has to be removed. This includes for example the projected seats that correspond to polling percentages or highlighted events that were relevant to the election cycle (e.g. the date of a political assassination). 
\item For some countries, the polling data are aggregated into a single .csv file. For other countries, data are broken down into multiple .csv files often in unpredictable ways. For example in some cases multiple .csv files exist, corresponding to different years since the last election. I merged the different .csv files where applicable.
\end{itemize}

\begin{figure}[t!]
     \centering
     \begin{subfigure}[b]{0.3\textwidth}
         \centering
         \includegraphics[width=\textwidth]{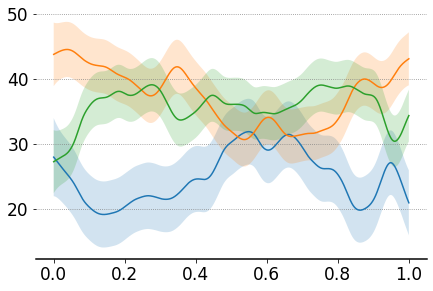}
         \caption{UK, 1987 election}
         \label{fig:uk_line}
     \end{subfigure}
     \hfill
     \begin{subfigure}[b]{0.3\textwidth}
         \centering
         \includegraphics[width=\textwidth]{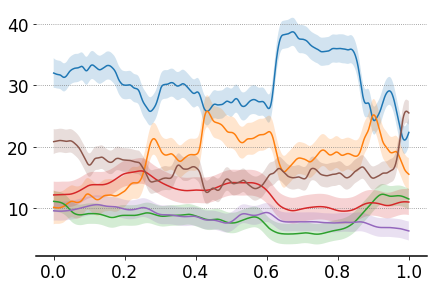}
         \caption{Germany, 2021 election}
         \label{fig:germany_line}
     \end{subfigure}
     \hfill
     \begin{subfigure}[b]{0.3\textwidth}
         \centering
         \includegraphics[width=\textwidth]{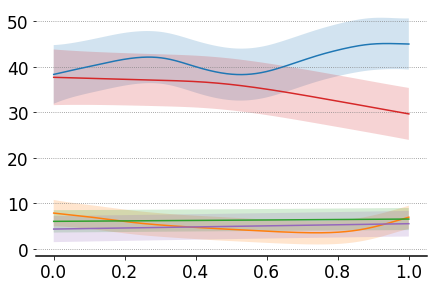}
         \caption{Portugal, 2005 election}
         \label{fig:portugal_line}
     \end{subfigure}
        \caption{Inferred time-series for different parties, for the UK 1987 election, the German 2021 election, and the Portuguese 2005. The confidence implied by the Gaussian process regressor varies widely by country.}
        \label{fig:three_europepolls_graphs}
\end{figure}

In summary, in order to create the Europepolls dataset the following preprocessing steps were applied.
\begin{enumerate}
\item I removed spurious columns so as to standardize the column types across countries.
\item I modified the column names where applicable so as to make them standardized.
\item I removed columns corresponding to Undecided Voters/ Blank Votes and renormalized the rest of the columns such that they summed up to 100\% of the vote.
\item I cleaned and standardized the dates, sample size, polling firm and commissioner columns.
\item I split coalition parties into their constituent parties. 
\item I removed the final election results where applicable. The preprocessed csv files contain only polling data and no election data.
\end{enumerate}

The preprocessing script used to preprocess the raw data is included in the supplementary material and is also available in the relevant GitHub repository.

\tocless{\section{Dataset Overview}}
The dataset includes opinion polls from 1983 to 2022 with an aim to include all EU members as well as Switzerland and the UK. The countries for which no data are currently provided are Luxembourg, the Netherlands, and Switzerland. This is mainly due to poor raw data quality at the time that the data was collected. In addition, data for Belgium is of poor quality but included in the dataset. Entries for Poland start in 1991, the UK in 1983, and Spain in 1986. However,  the mean starting date for all other countries is 2007. A detailed table of the starting and ending date for all countries can be found in Table \ref{table-start-end-date}

\tocless{\section{Experiments using the Europepolls dataset}}

\tocless{\subsection{Deriving a time-series from Europepolls data}}
One proposed use case for the Europepolls dataset is to infer a time series from the available polls,  which can then be used for further analyses. One key problem when inferring a trendline is trading bias and variance. A model that is too flexible will fit all the data perfectly (including noise). A model that is too inflexible will provide a poor fit. We propose to use a Gaussian Process regressor for inferring a trendline, as this regressor automatically trades off bias and variance based on marginal likelihood optimization. We use the scikit-learn package \citep{scikit-learn} and a Gaussian process regressor with a Matérn kernel with smoothness parameter $\nu=0.5$. We plot in Figure \ref{fig:three_europepolls_graphs} the inferred trendlines for the UK 1987 election, the German 2021 election and the Portuguese 2005 election. These trendlines can be used for inferring statistical relationships between complex socio-economic events and voting behavior.
\tocless{\subsection{Measuring pollster bias}}
We observe that the Gaussian process regressor fits the data by assuming a relatively large noise variance. Both for the UK and the Portuguese elections, the confidence interval is $\approx \pm 5\%$ larger than is typically implied by the statistical error of opinion polling. One possible source of these confidence intervals is a systematic bias exhibited by some pollsters (something that is known as ``house effects"). This could signal either intentional manipulation of the polling results or a flawed methodology. We measure the bias of different pollsters by calculating the mean error of each pollster from the inferred trendline. If the pollster is unbiased we should observe close to 0 error as the different errors would cancel out. We plot the results for such an evaluation in Table \ref{table-bias}. We observe that although some pollsters have a reasonable error with respect to the trendline others deviate significantly and consistently from the inferred mean. This analysis can be used to debias pollsters, reduce the noise in the trendline, and then explore other more involved analyses.

\begin{table}[!t]
  \caption{Pollster bias for the Portuguese 2005 election. We observe that although some pollsters have a reasonable error with respect to the trendline (e.g. Eurosondagem) others deviate significantly and consistently from the inferred mean (e.g. TNS). This analysis can be used to debias pollsters and reduce the noise in the trendline. One can then explore other more involved analyses, such as LLMs and ConvNets.}
  \label{table-bias}
  \centering
  \begin{tabular}{llllll}
    \toprule
 Country        & PS & PSD & BE & CDU & CDS-PP \\    
\midrule
 Aximage        & $-1.22$       & $-1.55$       & $-1.16$       & $-0.57$       & $+0.24$       \\
 Marktest       & $+1.02$       & $+1.10$       & $+0.66$       & $+0.41$       & $-0.25$       \\
 Eurosondagem   & $-0.50$       & $+0.51$       & $-0.30$       & $-0.17$       & $+0.78$       \\
 IPOM           & $+2.51$      & $+0.70$       & $+0.54$       & $-0.55$       & $+1.19$       \\
 Intercampus    & $+1.21$       & $+1.15$       & $-0.61$       & $+1.30$       & $-0.84$       \\
 TNS            & $-5.41$       & $+0.18$       & $+0.05$       & $-1.55$       & $-0.21$       \\
 Universidade Católica        & $+2.11$       & $-0.90$       & $+1.92$       & $+0.88$       & $-1.15$       \\
  \bottomrule
  \end{tabular}
\end{table}

\tocless{\section{Known weaknesses}}
The final dataset has a number of known weaknesses. 
\begin{itemize}
    \item The most important weakness is that the raw data has been scraped from Wikipedia. As such, it is difficult to assert if it has been tampered with. The quality of the information in general might be problematic as it is difficult to assess the skill, honesty, and carefulness of individual Wikipedia contributors. While I checked some sampled entries and found them to be accurate it was infeasible to check all the entries for correctness.
    \item Coalitions have been difficult to deal with automatically. Results might be not reflective of the political reality, or particularly informative.
    \item Some countries (e.g. Latvia) currently contain little or no data. This is often the result of poor raw data. 
    \item Other data sources. By restricting myself to data available on Wikipedia it is highly likely that I missed other publicly available data. A particular concern is data hiding behind language barriers, for example in local websites.
\end{itemize}

\tocless{\section{Safety and Ethical Discussion}}
The Europepolls dataset was constructed from opinion polling data already publicly available in aggregate form on Wikipedia. At the same time, the EuropeElects website publishes its own aggregate polling data. This however doesn't preclude the possibility that individual pollsters might consider their polling numbers as intellectual property, especially in aggregate form. I did not have the resources to investigate the legal framework of each individual European country surrounding the usage of opinion polls. Instead, I had to trust the due diligence done by EuropeElects before releasing their own dataset. EuropeElects is however a well-known entity in the European Union with collaborations with various international media organizations. As such I can be very confident, though not completely sure, that the collection, preprocessing, and dissemination of this dataset, is both legal and ethical. As a concluding note, I have provided a statement where I bear all responsibility in case of violation of rights and am both willing and able to remove any requested entries.

The use of machine learning for election forecasting has a number of ethical pitfalls. It can be argued that when data and its analysis are not open, machine learning can be used for the \emph{manipulation} of an election as opposed to contributing to a pre-election democratic discussion. The lines are often blurred between machine learning approaches that inform a party's campaign such that it better aligns with voters' needs, and machine learning approaches that give an unfair advantage to a campaign such that it manipulates voters. Another important ethical issue that we cannot fully address is that even if data and its analysis is publicly available, often the general public does not have the necessary skills to understand and contextualize the results. This creates a sense of mistrust towards machine learning and statistical approaches to social issues, where these approaches seem to obfuscate rather than elucidate social reality. I can only comment that this is one of the reasons that I chose a non-commercial license for this dataset. I also believe that ultimately open data will contribute positively to the democratic process.

\tocless{\section{Discussion}}
A number of interesting problems remain open. For many countries, the results are far from complete. It would be interesting then to look for polling data from different sources so as to i) increase the dataset ii) reduce possible errors. 

\bibliography{neurips_data_2023}
\bibliographystyle{abbrvnat}

\clearpage

\appendix
\section*{Appendix}
\tableofcontents

\section{Checklist}

The checklist follows the references.  Please
read the checklist guidelines carefully for information on how to answer these
questions.  For each question, change the default \answerTODO{} to \answerYes{},
\answerNo{}, or \answerNA{}.  You are strongly encouraged to include a {\bf
justification to your answer}, either by referencing the appropriate section of
your paper or providing a brief inline description.  For example:
\begin{itemize}
  \item Did you include the license to the code and datasets? \answerYes{See Section }
  \item Did you include the license to the code and datasets? \answerNo{The code and the data are proprietary.}
  \item Did you include the license to the code and datasets? \answerNA{}
\end{itemize}
Please do not modify the questions and only use the provided macros for your
answers.  Note that the Checklist section does not count towards the page
limit.  In your paper, please delete this instructions block and only keep the
Checklist section heading above along with the questions/answers below.

\begin{enumerate}

\item For all authors...
\begin{enumerate}
  \item Do the main claims made in the abstract and introduction accurately reflect the paper's contributions and scope?
    \answerYes{}
  \item Did you describe the limitations of your work?
    \answerYes{See Section 8 "Known weaknesses" in the main text.}
  \item Did you discuss any potential negative societal impacts of your work?
    \answerYes{See Section 9 "Safety and Ethical Discussion" in the main text.}
  \item Have you read the ethics review guidelines and ensured that your paper conforms to them?
    \answerYes{}
\end{enumerate}

\item If you are including theoretical results...
\begin{enumerate}
  \item Did you state the full set of assumptions of all theoretical results?
    \answerNA{}
	\item Did you include complete proofs of all theoretical results?
    \answerNA{}
\end{enumerate}

\item If you ran experiments (e.g. for benchmarks)...
\begin{enumerate}
  \item Did you include the code, data, and instructions needed to reproduce the main experimental results (either in the supplemental material or as a URL)?
    \answerYes{I provided a URL to the dataset which is hosted on a GitHub repository. I am also providing some additional code that was required to create the figures of the main text with the supplementary material.}
  \item Did you specify all the training details (e.g., data splits, hyperparameters, how they were chosen)?
    \answerYes{The main text contains some simple regression experiments using Gaussian processes. I specify the hyperparameters that I chose, together with the package that I used.}
	\item Did you report error bars (e.g., with respect to the random seed after running experiments multiple times)?
    \answerYes{The regression plots in the main text contain error bars that are the result of the Gaussian process regression. A discussion of their cause is also done in the main text.} 
	\item Did you include the total amount of compute and the type of resources used (e.g., type of GPUs, internal cluster, or cloud provider)?
    \answerNA{The dataset in it's current form was very cheap to preprocess requiring only a few minutes on the CPU of a MacBookPro2015}
\end{enumerate}

\item If you are using existing assets (e.g., code, data, models) or curating/releasing new assets...
\begin{enumerate}
  \item If your work uses existing assets, did you cite the creators?
    \answerYes{I'm building on top of curating work done by individual Wikipedia contributors. I'm citing Wikipedia as a whole to give credit to this work. I cite software packages that I used where applicable.}
  \item Did you mention the license of the assets?
    \answerYes{I include the license under which Wikipedia is licensed in Section \ref{assets_licence_section} of the Appendix.}
  \item Did you include any new assets either in the supplemental material or as a URL?
    \answerNo{}
  \item Did you discuss whether and how consent was obtained from people whose data you're using/curating?
    \answerNo{I discuss possible licensing issues that could exist with curating the results of multiple pollsters in Section 9 "Safety and Ethical Discussion" of the main text.}
  \item Did you discuss whether the data you are using/curating contains personally identifiable information or offensive content?
    \answerNA{The data is aggregate opinion polling data and as such does not contain identifiable or offensive content.}
\end{enumerate}

\item If you used crowdsourcing or conducted research with human subjects...
\begin{enumerate}
  \item Did you include the full text of instructions given to participants and screenshots, if applicable?
    \answerNA{}
  \item Did you describe any potential participant risks, with links to Institutional Review Board (IRB) approvals, if applicable?
    \answerNA{}
  \item Did you include the estimated hourly wage paid to participants and the total amount spent on participant compensation?
    \answerNA{}
\end{enumerate}

\end{enumerate}

\section{Data documentation and intended uses}
Here I follow the datasheets for datasets \citep{gebru2021datasheets} format so as to provide structured documentation for the proposed dataset that covers a number of common dataset parameters and points of concern.

\subsection{Motivation}
\begin{itemize}
\item \textbf{For what purpose was the dataset created?} The primary motivation behind the creation of this dataset was to spur modern machine learning research on the task of election and opinion poll forecasting. Up to now election forecasting has been primarily been conducted through opinion polling, prediction markets, and simple regressions from aggregate economic data. I firmly believe that modern deep learning approaches (Convolutional Networks, LLMs) can tackle the multi-modal data relevant in an election cycle so as to better forecast both elections and swings in opinion polls. My own personal motivation is that this would lead to better governance by better aligning government action with the opinion of voters.
\item \textbf{Who created the dataset?} The aggregation and curation of this dataset was done only by myself (Pitas Konstantinos). My work was based on previous work done by anonymous contributors to the Wikipedia online encyclopedia.
\item \textbf{Who funded the creation of the dataset?} I funded the creation of the dataset by myself and did not receive any outside funding. I created the dataset during a career brake that I took after my PhD defence.
\end{itemize}

\subsection{Composition}
\begin{itemize}
\item \textbf{What do the instances that comprise the dataset represent?} Each entry in the dataset represents that opinion polling measurement by a pollster, for a specific country and for a specific time period. An opinion polling measurement is the voting intention percentage for each political party large enough to be represented by name in the poll. More specifically a percentage of the vote is assigned to each party and this percentage is over all valid votes. This excludes votes that would have been invalid, and it also excludes responders to the poll that refused to answer to the pollster's questions.
\item \textbf{Does the dataset contain all possible instances or is it a sample of instances from a larger set?} The dataset is a non-random sample from a larger set. Specifically, it is currently unknown to me exactly how much polling data is in principle available, for which countries and for which time periods. European countries have a varied political history and have exhibited varied technological development. As such, autocratic periods mean that no polls where conducted for certain countries, while some less developed countries have only recently began to conduct polls. The current dataset is restricted to opinion polling data that was available \emph{on the English version of Wikipedia in 2021}.
\item \textbf{What does each instance consist of?} Each instance of the dataset has the following entries Date: The date when the poll was conducted, Polling Firm: The name of the firm conducting the poll, Commissioner: The name of the public or private entity which commissioned the poll, Sample Size: The size of the sample of the poll, Named Parties: All parties with high enough polling numbers such that they are included by name in the poll, Others: This entry includes the polling total of all parties polling too low to be included by name in the results.
\item \textbf{Is there a label or target associated with each instance?} The opinion polling data for each election cycle are targeting an election. The final results of this election can be thought as a label. However, I do not provide these final results of each election. These final results have to be retrieved from some other dataset.
\item \textbf{Is any information missing from individual instances?} Various entries are missing for different instances. The most common missing entries are the Commissioner field, the Sample Size field, and the Others field. This is especially the case for older polling data.

\item \textbf{Are relationships between individual instances made explicit?} The opinion polling data are organized in folders according first to the country where they were conducted and then for which year they were conducted. As such some high level logical relationships are made explicit by the dataset structure.

\item \textbf{Are there recommended data splits?} Polling data is inherently not independent. Within the same time series current polling instances depend statistically on previous polling instances. Between election cycles newer cycles are not independent of old cycles. The above have to be taken into consideration when constructing dataset splits. Dataset splits into training, testing and validation sets typically assume independent and identically distributed data points.

\item \textbf{Are there any errors, sources of noise, or redundancies in the dataset?} Yes. The data was gathered from Wikipedia. Various anonymous volunteers contributed to the Wikipedia data. It is hard to judge the honesty, integrity, and skill of these volunteers, who could have both done mistakes when reporting the data and/or have manipulated the data on purpose. Apart from the above, the data was gathered and curated by a single person without redundancies for error correction. By their very nature opinion polls are very noisy and suffer from not only from sampling noise, but also a number of potential methodological errors. Finally opinion polls can be a poor predictor of final election results especially as one is further from the election.

\item \textbf{Is the dataset self-contained, or does it link to or otherwise rely on external resources?} The dataset is self-contained.

\item \textbf{Does the dataset contain data that might be considered confidential?} No the dataset does not contain confidential data.

\item \textbf{Does the dataset contain data that, if viewed directly, might be offensive, insulting, threatening, or might otherwise cause anxiety?} No the dataset does not contain offensive content
\end{itemize}

\subsection{Collection Process}
\begin{itemize}
\item \textbf{How was the data associated with each instance acquired?} Each instance was gathered from wikipedia using the \href{https://wikitable2csv.ggor.de/}{wikitable2csv} tool.
\item \textbf{ If the dataset is a sample from a larger set, what was the sampling strategy?} The dataset is a sample from the larger set of all opinion polls conducted in the countries forming the European Union in 2021 (as well as the UK and Switzerland) in the modern period. The subset is the one that was available publicly on Wikipedia in 2021.

\item \textbf{Who was involved in the data collection process (e.g., students, crowdworkers, contractors) and how were they compensated (e.g., how much were crowdworkers paid)?} I was the only one involved in the dataset collection. I self-funded the creation of the dataset.

\item \textbf{Over what timeframe was the data collected?} The dataset was collected from January 2021 to December 2021.

\item \textbf{Were any ethical review processes conducted (e.g., by an institutional review board)?} No ethical review process was conducted. 
\end{itemize}

\subsection{Preprocessing/cleaning/labeling}
\begin{itemize}
\item \textbf{Was any preprocessing/cleaning/labeling of the data done (e.g., discretization or bucketing, tokenization, part-of-speech tagging, SIFT feature extraction, removal of instances, processing of miss- ing values)?} Yes, a number of preprocessing steps were applied to the dataset. Very broadly: I removed spurious columns so as to standardize the column types across countries, I modified the column names where applicable so as to make them standardized, I removed columns corresponding to Undecided Voters/ Blank Votes and renormalized the rest of the columns such that they summed up to 100\% of the vote, I cleaned and standardized the dates, sample size, polling firm and commissioner columns, I split coalition parties into their constituent parties, I removed the final election results where applicable. The preprocessed csv files contain only polling data and no election data. For more details one can consult the preprocessing scripts found in "europepolls/utils/preprocess\_utils.py".
\item \textbf{Was the “raw” data saved in addition to the preprocessed/cleaned/labeled data?} Yes the folder of each country contains a "data" folder which contains the raw .csv files gathered from Wikipedia.
\item \textbf{Is the software that was used to preprocess/clean/label the data available?} Yes the scripts used to preprocess the data are available in the "utils" folder of the dataset.  
\end{itemize}
\section{Uses}
\begin{itemize}
\item \textbf{Has the dataset been used for any tasks already?} The dataset has already been used by me to conduct statistical analyses in the context of the 2019-2023 Greek election cycle. Results are available on \href{https://www.kassiope.org/}{kassiope.org} (only in Greek).
\item \textbf{Is there a repository that links to any or all papers or systems that use the dataset?} No there is currently no need for such a repository.
\item \textbf{What tasks could the dataset be used for?} My intention is for the polling data to be used to infer a timeseries quantifying the fluctuating influence of each political party. This timeseries could then be used as the target of complex deep learning models which have multi-modal data as inputs (e.g. news articles). The practical utility of such models could be to better align public policy with voter intentions. Thus if a decline in economic variables will in the future result in loss of public confidence a government can act preemptively. Other uses could be found by financial companies aiming to hedge against political risk.
\item \textbf{Is there anything about the composition of the dataset or the way it was collected and preprocessed/cleaned/labeled that might impact future uses?} The dataset could be seen as too noisy or untrustworthy to be used in governance or finance as it was collected through Wikipedia.
\item \textbf{Are there tasks for which the dataset should not be used?} The dataset could in principle be used to manipulate elections. This could be for example by identifying events that significantly impact opinion polls and magnifying them through propaganda. I strongly discourage such use of this dataset.
\end{itemize}

\subsection{Distribution}
\begin{itemize}
\item \textbf{Will the dataset be distributed to third parties outside of the en- tity (e.g., company, institution, organization) on behalf of which the dataset was created?} Yes the dataset is currently freely available on GitHub. The current licence is the CC BY-NC 4.0 licence.
\item \textbf{Have any third parties imposed IP-based or other restrictions on the data associated with the instances?} To the best of my knowledge no such restrictions exist.
\item \textbf{Do any export controls or other regulatory restrictions apply to the dataset or to individual instances?} To the best of my knowledge no such restrictions exist.
\end{itemize}

\subsection{Maintenance}
\begin{itemize}
\item \textbf{Who will be supporting/hosting/maintaining the dataset?} For the moment I am the sole supporter, host, and maintainer of the dataset. However, even though the dataset is large for its category it is quite small by modern deep learning standards.
\item \textbf{How can the owner/curator/manager of the dataset be contacted (e.g., email address)?} I can be best contacted either through my institutional email pitas.konstantinos@inria.fr, or my personal email cwstas2007@hotmail.com.
\item \textbf{Will the dataset be updated (e.g., to correct labeling errors, add new instances, delete instances)?} Yes I plan to update the dataset. I envision updates to be yearly. This is because more recent data is not strictly within the scope of this dataset (other researchers such as EuropeElects consistently collect and publish opinion polls from 2018 and later). I expect opinion polls going further into the past to be uncovered infrequently.
\item \textbf{Will older versions of the dataset continue to be supported/hosted/maintained?} I expect the current repository to be updated with older version being available though GitHub versioning.
\item \textbf{If others want to extend/augment/build on/contribute to the dataset, is there a mechanism for them to do so?} One of the goals of publishing this dataset is to attract other contributors. This can be done by contributing to the main GitHub repository.
\end{itemize}

\section{Format}
The polling data are organized by election year. Thus the folder europepolls/countries/Austria/ contains the files 2013-09-29\_to\_2017-10-15, 2017-10-15\_to\_2019-09-29, 2019-09-29\_to\_2023-09-29 which correspond to three electoral periods. Each electoral folder contains a preprocessed\_polls.csv file which contains the preprocessed data. The electoral folder also contains a ``data" subfolder which contains .csv files that correspond to the raw polls scraped from Wikipedia.

Each preprocessed\_polls.csv file has the following format.

\begin{center}
\begin{tabular}{ |c|c|c|c|c|c| } 
 \hline
 Date & Polling Firm & Commissioner & Sample Size & Named Parties & Others \\ 
 \hline
 pandas.datetime & str & str & float & float & float \\ 
 \hline
\end{tabular}
\end{center}

A brief explanation of each column is given below.

\begin{itemize}
\item \textbf{Date}: The date when the poll was conducted. In some cases both a start and an end date is reported in the raw data,
corresponding to the start and the end of the polling. In this case, we report the end date. This column has ``pandas.datetime" entries.
\item \textbf{Polling Firm}: The name of the firm conducting the poll. This column has ``str" entries.
\item \textbf{Commissioner}: The name of the public or private entity which commissioned the poll. This column has ``str" entries.
\item \textbf{Sample Size}: The size of the sample of the poll. This column has ``float" entries.
\item \textbf{Named Parties}: The polling percentage of all parties with high enough polling numbers such that they are included by name in the poll. This column has ``float" entries ranging from 0-100\%. 
\item \textbf{Others}: This entry includes the polling total of all parties polling too low to be included by name in the results.
\end{itemize}

\section{Usage Example}
First, assume that we define the following functions that fit a Gaussian process to the opinion polling data of each party.
\begin{lstlisting}[language=Python, caption=Defining some preprocessing functions.]
from sklearn.gaussian_process import GaussianProcessRegressor
from sklearn.gaussian_process.kernels import RBF, WhiteKernel, Matern
import numpy as np
import pandas as pd
    
def get_regression(df_partial, party, start_date=None, end_date=None):
    """
    Fits a Gaussian Process on the polling data of a single party. The difference with fit_gp is that regressed values
    are returned for every day withing the specified range, as opposed to only returning values when polls are
    available.

    Parameters
    ----------
    df_partial: pandas.dataframe
        A pandas dataframe containing the polling data for the different political parties. The dataframe structure has
        to be Date | Polling Firm | Commissioner | Sample Size | {Party Names} | Others.
    party: str
        The name of the party for which to fit the GP.
    start_date: str
        The starting date to compute the regression. Should be of the form 'YYYY-MM-DD'.
    end_date: str
        The ending date to compute the regression. Should be of the form 'YYYY-MM-DD'.

    Returns
    -------
    d: pandas.dataframe
        A pandas dataframe with two columns party_mean | party_std containing the regressed mean and standard deviation
        respectively.
    """

    df_partial['Date'] = pd.to_datetime(df_partial['Date'])
    df_partial['Date_delta'] = (df_partial['Date'] - df_partial['Date'].min()) / np.timedelta64(1, 'D')
    norm = df_partial['Date_delta'].max()
    df_partial['Date_delta'] = df_partial['Date_delta']/norm

    if start_date is None and end_date is None:
        detailed_dates = pd.date_range(df_partial['Date'].min(), df_partial['Date'].max(), freq='d')
    else:
        detailed_dates = pd.date_range(start_date, end_date, freq='d')
    normalized_detailed_dates = ((detailed_dates - df_partial['Date'].min()) / np.timedelta64(1, 'D'))/norm

    df_partial = df_partial.dropna()

    kernel = Matern() + WhiteKernel()
    gpr = GaussianProcessRegressor(kernel=kernel, random_state=0).fit(df_partial[['Date_delta']], df_partial[party])

    pred_mean, pred_std = gpr.predict(normalized_detailed_dates.to_numpy().reshape(-1, 1), return_std=True)

    d = {
        'Date': detailed_dates,
        party + '_mean': pred_mean,
        party + '_std': pred_std
    }

    return pd.DataFrame(d)

def compute_regression(df, start_date=None, end_date=None):
    """
    Computes a regression for all days within the specified time interval and for all available parties.

    Parameters
    ----------
    df: pandas.dataframe
        A pandas dataframe containing the polling data for the different political parties. The dataframe structure has
        to be Date | Polling Firm | Commissioner | Sample Size | {Party Names} | Others.
    start_date: str
        The starting date to compute the regression. Should be of the form 'YYYY-MM-DD'.
    end_date: str
        The ending date to compute the regression. Should be of the form 'YYYY-MM-DD'.

    Returns
    -------
    regressed: pandas.dataframe
        A pandas dataframe with columns {party}_mean | {party}_std containing the regressed mean and standard deviation
        respectively for all days within the specified interval, and for all available parties.
    """

    all_parties = set(df.columns) - set(['Date', 'Polling Firm', 'Commissioner', 'Sample Size', 'Others'])

    regressions = []

    for party in all_parties:

        df[party] = df[party]/100

        regressions.append(get_regression(df[[party, 'Date']], party=party, start_date=start_date, end_date=end_date))

    regressions = [df.set_index('Date') for df in regressions]
    regressed = pd.concat(regressions, axis=1)*100
    regressed.reset_index(inplace=True)
    regressed = regressed.rename(columns={'index': 'Date'})
    return regressed
\end{lstlisting}
Assume also that the polling data of the UK 1987 election exists in the file 'uk\_1983-06-09\_to\_1987-06-11.csv'. Then to fit a Gaussian process on the opinion polling data of each party participating in the election, one simply needs to execute the following instructions. 
\begin{lstlisting}[language=Python, caption=Loading and preprocessing the opinion polling data for the UK 1987 election.]
uk = pd.read_csv('uk_1983-06-09_to_1987-06-11.csv')
df_reg_uk = compute_regression(uk)
\end{lstlisting}

\section{Author Statement}
I, Pitas Konstantinos, hereby provide this author statement for the dataset titled Europepolls, which I have submitted to NeurIPS 2023. I affirm that I am the creator and sole owner of this dataset, and I bear full responsibility for any potential violations of rights, legal issues, or ethical concerns that may arise from its usage.

By submitting this dataset to NeurIPS 2023, I acknowledge and confirm that I have chosen the "Creative Commons Attribution-NonCommercial 4.0 International" data license for this dataset. This license permits others to use, adapt, and distribute the dataset for non-commercial purposes while requiring appropriate attribution to the original creator, which is myself.

I understand that NeurIPS has certain guidelines and policies regarding the submission and handling of datasets. I have made every effort to ensure the dataset's accuracy, integrity, and compliance with applicable laws, regulations, and ethical standards. I furthermore state that NeurIPS and its organizers bear no responsibility for any issues that may arise from the usage of this dataset.

I affirm that this dataset does not violate any intellectual property rights, copyrights, or any other legal or ethical obligations. In the event that any claims or disputes arise in connection with the dataset, I will assume full responsibility.

I declare that I have read, understood, and agreed to the above statements and that all the information provided is accurate and truthful to the best of my knowledge.

\section{Hosting, licensing and maintenance plan.}
\textbf{Hosting}
For the moment I am planning to continue hosting the dataset on my personal GitHub \href{https://github.com/konstantinos-p/europepolls}{page}. In the near future, I am planning to submit the dataset on other platforms and dataset repositories, such that the dataset remains accessible in the unforeseen event where my GitHub page becomes inaccessible. A specific platform that I aim to submit my dataset to, is the Kaggle website. I believe the proposed dataset will serve a very good educational role on Kaggle as for many machine learning/data-scientist practitioners it is one of the first points of contact with real data. Other possible repositories include \href{https://figshare.com/}{figshare}, \href{https://data.mendeley.com/}{Mendeley Data}, and \href{https://dataverse.harvard.edu/}{Harvard Dataverse}.

\textbf{Licencing}
Concerning the licensing of the dataset, I aim to continue to provide the dataset with it's current CC BY-NC 4.0 licence.

\textbf{Maintenance}
While the dataset captures a large part of the opinion polling data available on Wikipedia, it is possible that it is not exhaustive. New articles and data are added to Wikipedia daily and it is highly likely that more opinion polling data going back further into the past will become available. I plan to continue monitoring any such data that become available and add them to the existing data. One main aim of making this dataset public is allowing contributors from the highly multilingual European Union to contribute to the improvement of the dataset. I believe that it is highly likely that more detailed data is available for certain countries but is currently inaccessible due to language barriers. At the same time, I plan to investigate other websites apart from Wikipedia that might contain opinion polling data that is not in the current dataset. Finally, I consider data after 2018 to be fairly easy to gather. For the moment, the website \href{https://europeelects.eu/}{EuropeElects} gathers opinion polling data consistently using volunteers and posts the collected data in easy-to-access csv files on a public repository. As such it is fairly straightforward to augment the existing dataset with opinion polling data for the years after 2018.

\section{URL to hosting website}
The dataset is available to download from GitHub using the following \href{https://github.com/konstantinos-p/europepolls}{link}.

\section{Persistent Dereferenceable Identifier}
A DOI generated through Zenodo and pointing to the latest release of the dataset can be found \href{https://zenodo.org/badge/latestdoi/441550470}{here}.

\section{Assets Licence}\label{assets_licence_section}
Wikipedia is the sole asset of this dataset. Wikipedia is licensed under the  Attribution-ShareAlike 3.0 Unported (CC BY-SA 3.0) licence.

A link to the license can be found \href{https://creativecommons.org/licenses/by-sa/3.0/}{here}.

\section{Licence}\label{licence_section}
The dataset has been released on GitHub under the "Creative Commons Attribution-NonCommercial 4.0 International" licence. Please note that this licence might be incompatible with the Wikipedia licence. I would be happy to discuss with the reviewers this point and make any necessary changes.

A link to the license can be found \href{https://creativecommons.org/licenses/by-nc/4.0/}{here}.

\section{Structured Metadata}
In it's current form the dataset doesn't required structured metadata as it is published on GitHub which relies on GitHub topic tags to facilitate search queries. However, I include a structured metadata json example for later use.

\begin{lstlisting}[language=HTML, caption=Structured metadata json file.]
<html>
  <head>
    <title>NCDC Storm Events Database</title>
    <script type="application/ld+json">
    {
      "@context":"https://schema.org/",
      "@type":"Dataset",
      "name":"NCDC Storm Events Database",
      "description":"A dataset of country-level historical voting intention polling data from European Union countries.",
      "url":"https://github.com/konstantinos-p/europepolls",
      "sameAs":"https://github.com/konstantinos-p/europepolls",
      "identifier": ["",
                     ""],
      "keywords":[
         "voting intention > opinion polling > elections > european parliament",
      ],
      "license" : "https://creativecommons.org/licenses/by-nc/4.0/",
      "isAccessibleForFree" : true,
      "hasPart" : [
      ],
      "creator":{
         "@type":"-",
         "url": "https://www.konstantinos-pitas.com/",
         "name":"Pitas Konstantinos, PhD",
         "contactPoint":{
            "@type":"ContactPoint",
            "contactType": "personal details",
            "telephone":"-",
            "email":"cwstas2007@hotmail.com"
         }
      },
      "funder":{
         "@type": "individual",
         "sameAs": "https://www.konstantinos-pitas.com/",
         "name": "Pitas Konstantinos, PhD"
      },
      "includedInDataCatalog":{
      },
      "distribution":[
         {
            "@type":"DataDownload",
            "encodingFormat":"CSV",
            "contentUrl":"https://github.com/konstantinos-p/europepolls"
         }
      ],
      "temporalCoverage":"1983-01-01/2022-12-31",
      "spatialCoverage": "European Union"
    }
    </script>
  </head>
  <body>
  </body>
</html>
\end{lstlisting}

\end{document}